\documentclass[conference]{IEEEtran}
\IEEEoverridecommandlockouts

\usepackage{cite}
\usepackage{amsmath,amssymb,amsfonts}
\usepackage{algorithmic}
\usepackage{graphicx}
\usepackage{textcomp}
\usepackage{xcolor}
\usepackage{hyperref}
\usepackage{booktabs}

\def\BibTeX{{\rm B\kern-.05em{\sc i\kern-.025em b}\kern-.08em
    T\kern-.1667em\lower.7ex\hbox{E}\kern-.125emX}}

\begin{document}

\title{Interventional Causal Circuits for Safe Robot Action Testing and Failure Recovery}

\author{
\IEEEauthorblockN{Naren Vasantakumaar, Tom Schierenbeck, Michael Beetz}
\IEEEauthorblockA{Institute for Artificial Intelligence, University of Bremen\\
\{naren, tom\_sch, beetz\}@uni-bremen.de}
}

\maketitle

\begin{abstract}
Safe physical AI for robot actions are required not only likely to succeed but tested to be safe before execution. In practice, however, formal testing of motion parameters is computationally expensive, and the cost scales poorly with the dimensionality of the action space. When a proposed action is rejected by a tester, the naive response is to resample blindly until a passing candidate is found. This is wasteful, uninformative, and offers no convergence. We argue that rejection should instead trigger causal diagnosis: a principled identification of which action parameter caused the failure and what corrective value maximises the probability of passing testing under the interventional probability distribution. We propose a closed-loop framework that couples a Joint Probability Tree (JPT)~\cite{joint_probability_trees} with a Causal Circuit derived from a Marginal-Deterministic Variable Tree~\cite{wang2023mdvtree}, enabling exact polytime computation without retraining, or additional data collection. The framework validates tractability of all interventional queries before the robot begins operating, and out-of-support candidates are detected and excluded from correction automatically. We perform experiments in a ROS2 simulation environment, and the framework demonstrates complementary roles across quality of distribution: under a high-quality JPT, the Causal Circuit reduces failed attempts by 10.3\% and under a degraded JPT, it reduces total failed attempts by 37\%. Every rejected plan produces a structured, interpretable causal report naming the primary cause variable, its observed value, and the recommended corrective region, supporting operator oversight and autonomous recovery without a separately trained failure model.
\end{abstract}


\section{Introduction}
\small

A central challenge in deploying physically safe autonomous systems~\cite{amodei2016concrete} is enabling robots to operate within a continuous loop of proposing actions, testing them against safety constraints, and adapting when those proposals fail. This iterative ``hypothesize, test, debug'' loop naturally arises in architectures where a planner generates candidate actions and a separate testing stage evaluates whether they satisfy constraints such as collision, reachability, stability, or learned safety criteria before execution, echoing the broader execution-monitoring and replanning literature in which discrepancies between expected and observed behavior trigger repair~\cite{pettersson2005monitoring,mendoza2015execution,beetz1994improving}. Such testing-centric pipelines are increasingly common in safe physical AI because they let planners remain flexible and expressive while delegating safety assessment to an external tester.

In robotics, before a robot arm reaches for an object or navigates through a constrained environment, the proposed motion parameters must first pass testing. Formal testing methods can provide strong guarantees, but they are computationally expensive, and the cost grows with both environmental complexity and the dimensionality of the action space~\cite{kressgazit2018synthesis,orthey2024review}. Since planners may need to submit many candidate actions before finding one that passes testing, the efficiency of this planning-testing loop becomes an important systems problem~\cite{recover2024,chitnis2019learning,garrett2021tamp}.

In many existing systems, a failed test simply triggers a new sampling step~\cite{ichter2018learning,kunz2016hierarchical}: the planner draws another candidate from the planning distribution and resubmits it, exploring the continuous action space without understanding every constraint explicitly. Yet each rejection also carries information about which parts of the action space are less compatible with safe execution, rather than being purely a failure signal. This creates an opportunity to extend the hypothesize-test loop with a structured debugging stage, enabling the robot not only to generate and evaluate plans but also to reason about and adapt to the causes of rejected proposals.

Existing recovery approaches address parts of this loop but often from different perspectives. Classical model-based methods detect symbolic precondition violations and replan from scratch~\cite{recover2024,sung2024learning}, discarding the continuous parametric structure of the rejected plan. Data-driven approaches, including reinforcement learning and vision-language pipelines~\cite{vifailback2025,ahmad2025unified,replan2024}, learn corrective behaviors from large offline datasets or online interaction. Our focus is different: we investigate whether causal reasoning can be integrated directly into the planning-testing loop itself, enabling robots to diagnose failed action hypotheses using only the fitted planning distribution and the tester outputs, without requiring a separately trained failure model.

Wang and Kwiatkowska~\cite{wang2023mdvtree} show that imposing a Marginal-Deterministic Variable Tree (MdVtree) on a probabilistic circuit enables exact polytime computation of interventional distributions via backdoor adjustment~\cite{pearl2009causality}. We build on this theoretical result by extending a Joint Probability Tree~\cite{joint_probability_trees}, already used for parameter sampling in robot action planning, into a \emph{Causal Circuit} satisfying these structural conditions without additional retraining or data collection.

This article makes two contributions. First, we present a practical construction that extends a trained probabilistic planning model (Joint Probability Tree) into a Causal Circuit by imposing an MdVtree, together with a support determinism verification step that validates tractability before deployment. Second, we present a closed-loop hypothesize-test-debug pipeline in which each rejected plan triggers causal diagnosis and a one-shot parameter correction on the identified cause variable, while all remaining parameters are resampled from the conditional planning distribution, preserving the correlations learned by the planner. Plans whose parameters lie entirely outside the training support are automatically detected and excluded from correction, ensuring that the system does not extrapolate corrective recommendations beyond observed data.

\section{Related Work}
\small
 
\textbf{Failure detection and recovery.}
Classical neuro-symbolic approaches detect execution failures by checking symbolic preconditions and respond by replanning from scratch~\cite{recover2024,sung2024learning}. These methods treat failure as a discrete, binary event and discard the continuous parametric variables of the rejected plan. When a motion fails because an approach position is geometrically infeasible by a small margin, replanning from scratch produces a new plan with no guarantee that the offending parameter is corrected. Data-driven approaches such as ViFailback~\cite{vifailback2025} and vision-language pipelines~\cite{ahmad2025unified} frame failure recovery as a perceptual context task, diagnosing failures from scene observations and generating corrective instructions. While expressive, these methods require large offline failure datasets.

\textbf{Causal inference in robotics.}
The use of causal models in robot learning has received growing attention~\cite{scholkopf2021causal,ahmed2021causalworld,lee2022crest}, primarily for policy generalisation and sim-to-real transfer. Pearl's do-calculus~\cite{pearl2009causality} and related causal frameworks~\cite{peters2017elements} provide the formal foundation for reasoning about interventions, and causal representations have been applied to structured world models~\cite{scholkopf2021causal} and transfer learning in manipulation~\cite{ahmed2021causalworld,lee2022crest}. Our work differs in that we do not use causal inference to train a better policy: the causal circuit operates at runtime on a fitted planning distribution for testing and diagnosing individual failed plans to increase safety and efficiency.

\textbf{Tractable probabilistic circuits.}
Probabilistic circuits~\cite{choi2020probabilistic} provide a unified framework for tractable probabilistic inference, supporting exact marginal, conditional, and joint queries in polytime. Wang and Kwiatkowska~\cite{wang2023mdvtree} extend this framework to causal inference, proving that marginal determinism enforced by the MdVtree structure enables exact, polytime computation of interventional distributions via backdoor adjustment. Our work experiments this theoretical framework in a robot planning setting, showing how a fitted JPT can be extended into a causally structured circuit and deployed with runtime testing and correction.

\textbf{Task and motion planning under uncertainty.}
The works~\cite{chitnis2019learning,garrett2021tamp} address the challenge of finding feasible continuous parameters for symbolic action sequences, and several works learn feasibility classifiers to reduce the number of motion planner calls. Our approach is complementary: rather than predicting feasibility before attempting a plan, we diagnose failure after rejection and produce a targeted correction, reducing the number of subsequent testing calls.
 
\section{System Overview}
\small
 
\begin{figure}[t]
\centering
\includegraphics[width=0.98\linewidth]{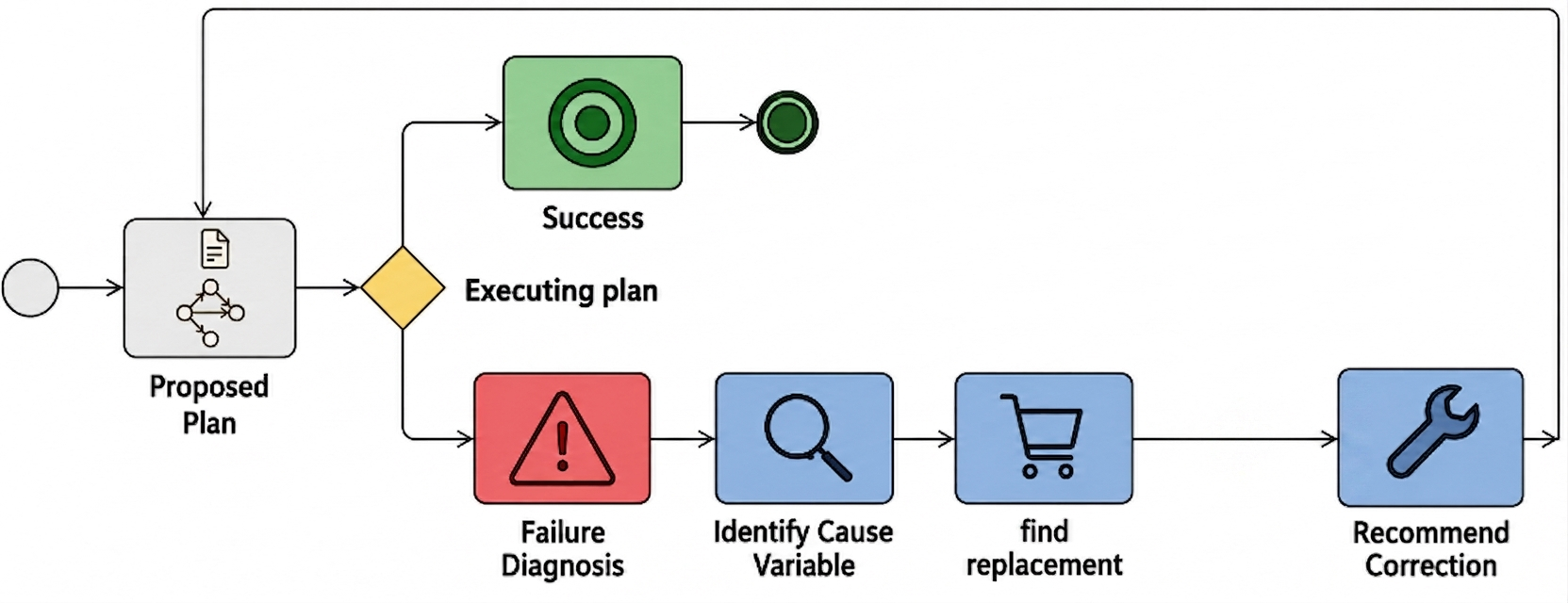}
\caption{The figure represents the proposed closed-loop framework using causal diagnosis. When a plan fails, the framework moves through diagnosis and identification to find a replacement and recommend a correction for the next attempt. }
\label{fig:system}
\end{figure}

The proposed framework is task-agnostic to any robot action that can be parameterised as a continuous vector $\mathbf{x}$ of action variables like approach positions, joint configurations, grasp parameters, weight of the object, or any combination thereof, and for which a tester can evaluate plan safety before execution. The only task-specific inputs are a set of successful training executions and a designation of which variables are causes and which constitute the effect. Figure~\ref{fig:system} shows the structure for this robot action testing framework.
 
Given a database of Narrative Enabled Episodic Memories (NEEMs)~\cite{beetz2025cram}, structured logs that record a robot plan's action parameters, context, and outcome for each execution, a Joint Probability Tree (JPT)~\cite{joint_probability_trees} is fitted to the successful executions. Like a regression tree, a JPT recursively partitions the parameter space into axis-aligned regions and fits a simple, independent distribution within each region; the resulting piecewise structure encodes the joint distribution $p(\mathbf{x})$ over all action parameters, captures inter-variable correlations, and supports fast, exact marginal and conditional inference. This fitted model serves as the planning distribution from which candidate actions are sampled at runtime.
 
To enable causal inference, we designate a set of cause variables $\mathbf{C} \subseteq \mathbf{x}$ and an effect variable $Y$. A causal priority ordering over $\mathbf{C}$ is established by normalised average treatment effect analysis~\cite{pearl2009causal,peters2017elements} (ATE$_\text{norm}$), which ranks each variable by its influence on $Y$ across the training set. This ordering is used to construct the Marginal-Deterministic Vtree (mdvtree)~\cite{wang2023mdvtree} over $\mathbf{C}$: a vtree is a binary tree whose leaves are the variables in $\mathbf{C}$ and whose internal nodes recursively split them into a nested left/right hierarchy, here ordered by ATE$_\text{norm}$ so that the most influential variable is separated first. Because a JPT already partitions its input space recursively like a decision tree, imposing this order and restricting splits to variables in $\mathbf{C}$ extends it into a \emph{Causal Circuit}: every SumUnit at each causal level is forced to partition its children's support disjointly on the corresponding cause variable, a property called \emph{marginal determinism}~\cite{wang2023mdvtree}.

This structural property, which the plain JPT does not possess, enables exact, polytime computation of interventional distributions via the backdoor adjustment formula~\cite{pearl2009causality}. A support determinism verification step validates this condition formally before deployment, providing a pre-runtime guarantee that all interventional queries will be tractable.
At runtime, the system operates in the closed loop shown in Figure~\ref{fig:system}. On each planning cycle, a candidate $\hat{\mathbf{x}}$ is sampled from the JPT and submitted to the tester. If the plan passes, the robot executes it and the loop resets. If the plan is rejected, rather than resampling blindly, the Causal Circuit diagnoses the rejection. For each cause variable $x_i \in \mathbf{C}$, the circuit evaluates the interventional success probability $\pi_i(v) = P(Y \geq \tau \mid do(x_i = v),\, \mathbf{C}_{-i} = \hat{\mathbf{x}}_{-i})$~\cite{pearl2009causality}, where $do(x_i = v)$ denotes Pearl's intervention operator for probability of success if we were to \emph{force} $x_i$ to take the value $v$, rather than merely observing it. The primary cause is the variable whose observed value minimises $\pi_i(\hat{x}_i)$; the recommended corrective value is $v^* = \arg\max_v\,\pi_i(v)$. Both are computed in closed form without any requirement for additional simulation, data collection, or retraining.
 
A \emph{Causal Sampling Correction} is then applied to the next planning cycle: the primary cause variable is constrained to the interval $[v^* - \delta,\, v^* + \delta]$, where $\delta$ is set to one JPT leaf width for the variable in question, while all other parameters are drawn from the conditional distribution $p(\mathbf{x}_{-i} \mid x_i{=}v^*)$. The correction is one-shot and non-chaining: if the corrected attempt succeeds, the system reverts to unconstrained JPT sampling; if it also fails, the correction is discarded rather than refined further, preventing the system from committing to a bad recommendation. Plans for which $\rho_i(\hat{x}_i) = 0$ for all $i$ indicates that the observed parameters lie entirely outside the training support are detected automatically and excluded from the correction loop. The ability to detect out-of-distribution inputs is a fundamental requirement for safety as the system has no reliable basis for predicting outcomes, and any correction is operating without safety guarantees.

\section{The Causal Circuit}
\small
 
The important technical system of the entire framework is the \emph{Causal Circuit}: a probabilistic circuit that extends a fitted JPT with the causal structure. This section describes how the circuit is constructed from the JPT, what structural conditions it must satisfy, how those conditions are tested before deployment, and how the circuit is queried during failure diagnosis.
 
\subsection{From Joint Probability Tree to Causal Circuit}
 
A JPT~\cite{joint_probability_trees} trained and fitted to successful executions encodes $p(\mathbf{x})$ as a probabilistic circuit~\cite{choi2020probabilistic} of \emph{SumUnits} (weighted mixtures) and \emph{ProductUnits} (factorisations), with univariate distributions at the leaves. A critical property for safety is JPTs' \emph{finite support}: each leaf covers a bounded axis-aligned interval, so $p(\mathbf{x}) = 0$ for any parameter combination absent from training~\cite{joint_probability_trees}. Unlike models that depend on Gaussian Distributions, which assign non-zero density everywhere, a JPT makes out-of-support detection exact rather than a soft approximation~\cite{joint_probability_trees}. However, a circuit fitted purely on observational data carries no causal semantics and cannot answer interventional queries $P(Y \mid do(x_i = v))$, the probability of $Y$ if we were to \emph{force} $x_i$ to a value, rather than merely observe it, because the learned distribution conflates the causal effect of $x_i$ with the influence of any confounders~\cite{pearl2009causality,pearl2009causal}.

\begin{figure}[t]
\centering
\includegraphics[width=0.98\linewidth]{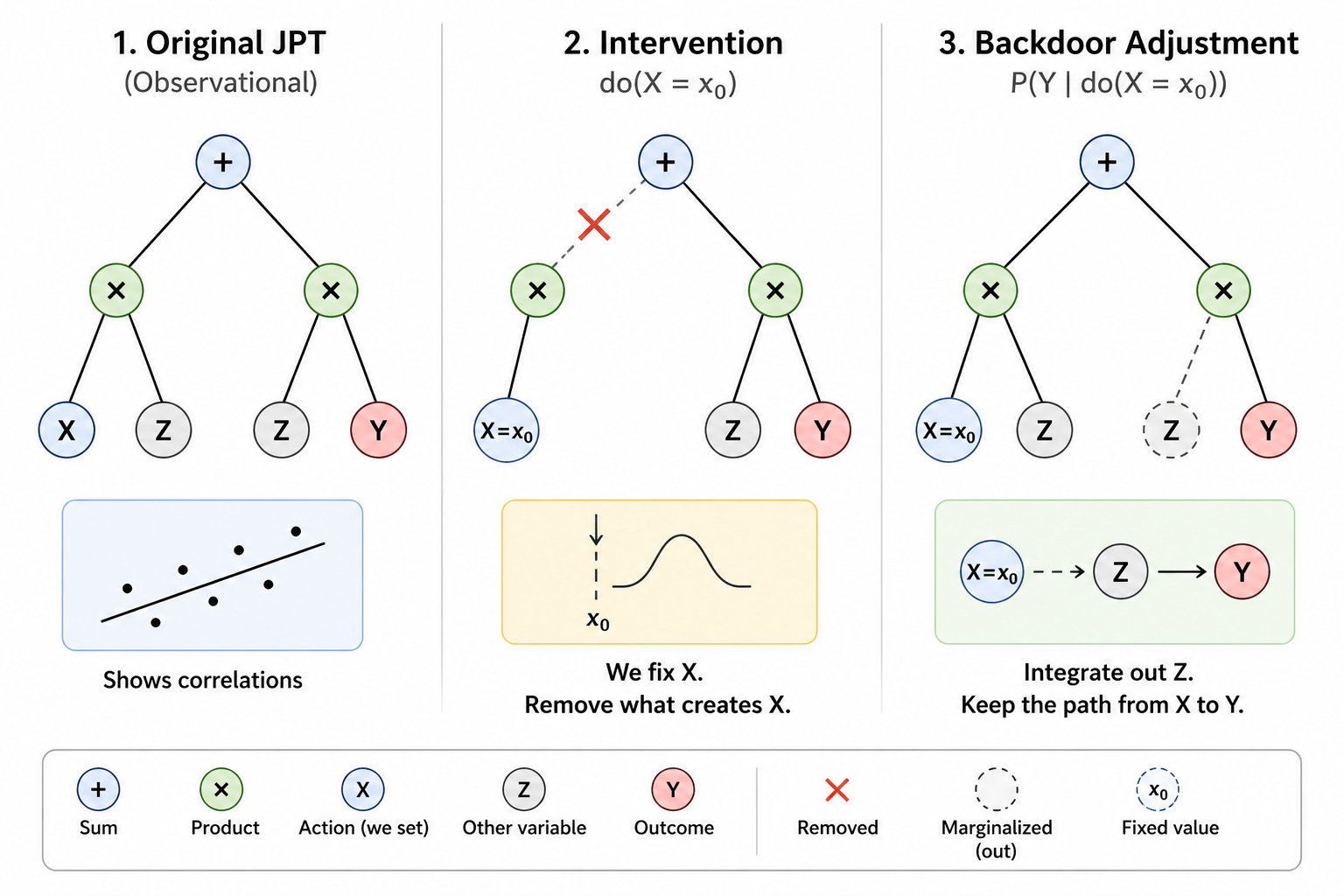}
\caption{Construction of the Interventional Causal Circuit. \textit{Left:} the original JPT encodes the observational joint distribution, \textit{Centre:} the intervention $do$ removes the incoming edges to the chosen cause variable $x_0$, \textit{Right:} the resulting interventional circuit applies backdoor adjustment by marginalising over the remaining cause variables $\mathbf{Z}$}
\label{fig:jptvscausal}
\end{figure}

Our contribution is the procedure connecting these components in a robot planning setting: we construct an MdVtree from the ATE$_\text{norm}$-ranked order over $\mathbf{C}$, placing the highest-priority variable at the root and recursively splitting the remainder, then verify the Q-determinism condition (support disjointness, defined formally in Sec.~\ref{sec:verification})~\cite{wang2023mdvtree,choi2020probabilistic} on the fitted JPT circuit to produce a \emph{Causal Circuit} without retraining. Figure~\ref{fig:jptvscausal} shows the fitting of a causal circuit with backdoor adjustment on a JPT.
 
\subsection{Support Determinism Verification}
\label{sec:verification}
 
Before the robot begins the task, the Causal Circuit runs a verification step that validates if the circuit satisfies marginal determinism for every declared cause variable.
 
The condition being checked is \emph{support disjointness}: for each $x_i \in \mathbf{C}$, every SumUnit that partitions on $x_i$ must have pairwise disjoint children supports on $x_i$ which is the Q-determinism condition of~\cite{wang2023mdvtree}. When it holds, the backdoor adjustment sum reduces to a weighted sum over non-overlapping leaf regions with no double-counting. When it fails, the same cause value is covered by multiple branches and the interventional probability is ill-defined. 

JPTs are well-suited to satisfying this condition: their axis-aligned splits naturally produce disjoint regions on each split variable, so marginal determinism tends to hold on the variables the JPT found most predictive, the same variables ranked highest by ATE$_\text{norm}$. When a violation is found, the result identifies the exact SumUnit and variable responsible. If verification passes, a \emph{Support Determinism Verification Result} is returned as a pre-deployment certificate that all interventional queries will be tractable and exact.

\subsection{Backdoor Adjustment and Interventional Circuits}
 
When a plan is rejected, the Causal Circuit computes the interventional distribution $P(Y \mid do(x_i = v))$ for each cause variable $x_i \in \mathbf{C}$ via the backdoor adjustment formula~\cite{pearl2009causality}:
\begin{align}
  &P(Y \geq \tau \mid do(x_i = v)) = \nonumber\\
  &\quad\sum_{z} P(Y \geq \tau \mid x_i = v,\, \mathbf{Z} = z)\; P(\mathbf{Z} = z),
  \label{eq:backdoor}
\end{align}
where $\mathbf{Z}$ is the adjustment set. The circuit computes this sum exactly and in polytime by exploiting the support determinism structure: because each SumUnit partitions the cause domain disjointly, the marginalisation over $x_i$ reduces to a weighted sum over non-overlapping leaf regions, with no double-counting~\cite{wang2023mdvtree}.
 
Concretely, in the JPT case, where the circuit's support decomposes into disjoint axis-aligned leaf regions, \emph{backdoor adjustment} constructs a new joint circuit over $(x_i, Y)$ as the \emph{interventional circuit}: as a SumUnit of ProductUnits, one per disjoint cause support region $R_k$ where,
\begin{align}
  &\text{SumUnit}\bigl[P(x_i \in R_k)\bigr] \;\to \nonumber\\
  &\quad\text{ProductUnit}\bigl[p(x_i \mid x_i \in R_k),\; p(Y \mid x_i \in R_k)\bigr].
  \label{eq:circuit}
\end{align}
Each leaf region $R_k$ corresponds to one axis-aligned interval in the JPT's partition of $x_i$'s domain. The cause branch is the truncated cause marginal over $R_k$; the effect branch is the truncated joint circuit conditioned on $x_i \in R_k$, then marginalised onto $Y$. All truncations and marginalisations are computed in closed form using the circuit's existing inference primitives, and no sampling or numerical integration is required. The resulting interventional circuit is tractable for the marginal and mode query classes, enabling both the probability evaluation and the corrective region extraction described below.
 
\subsection{Failure Diagnosis}
 
Given the observed parameter vector $\hat{\mathbf{x}}$ of a rejected plan, \emph{diagnose failure} produces the interventional circuit for each cause variable independently. For each $x_i \in \mathbf{C}$, the circuit evaluates how much probability mass the marginal interventional distribution places on a narrow interval around the observed value $\hat{x}_i$:
\begin{equation}
  \rho_i(\hat{x}_i) = \int_{\hat{x}_i - \epsilon}^{\hat{x}_i + \epsilon}
  p^{\,\text{do}}(x_i)\, dx_i,
  \label{eq:rho}
\end{equation}
where $p^{\,\text{do}}(x_i)$ is the marginal of the interventional distribution over $x_i$, and $\epsilon$ is a small query resolution matched to the JPT leaf precision. This gives a direct measure of how anomalous $\hat{x}_i$ is under the causal model: a low value of $\rho_i(\hat{x}_i)$ indicates that the interventional distribution assigns little mass near the observed value. The \emph{primary cause} is the variable with the lowest $\rho_i(\hat{x}_i)$: the one whose observed value the interventional distribution assigns the least plausibility.
 
Alongside the diagnosis, the circuit identifies the \emph{recommended region} $R^* = \arg\max_{R_k} P(x_i \in R_k \mid do(x_i))$, the cause support region with the highest interventional probability. The set $R^*$ is returned rather than a single point, so the downstream correction retains the bounds and can derive a corrective value from them directly. Plans for which $\rho_i(\hat{x}_i) = 0$ for all $i$ are flagged as out-of-support: the observed parameters lie entirely outside the training distribution, and no correction grounded in the training data is meaningful. These cases are excluded from the correction loop and reported as a distinct failure category.
 
The diagnosis is computed entirely from the trained circuit and no additional data, simulation, or learned failure model is required. Every rejected plan produces a structured \emph{Failure Diagnosis Result} that names the primary cause variable, reports its observed value and interventional probability, identifies the recommended corrective region and its probability, and lists the full per-variable profile across all cause variables. This structured output supports both autonomous correction and human-readable operator explanation.
 
\section{Experiment and Results}
\small
 
The framework is evaluated on a robot pick-and-place task of milk in a ROS2 simulation environment \footnote{Experiment code, training data, and model files are available at: \url{https://github.com/Narenvasant/causal_reasoning}}. A PR2 robot navigates to grasp a milk carton from a kitchen counter and places it on a dining table. The action parameter vector $\mathbf{x}$ comprises five variables: counter approach $(x, y)$, table approach $(x, y)$, and arm selection. The world geometry imposes a non-trivial constraint: only a subset of geometrically plausible approach positions permit a collision-free path through the narrow passage, and the JPT must guide sampling toward this feasible region without explicit knowledge of the deployment environment. 
 
A JPT is trained on 1,742 successful executions of the pick-and-place task in an open-world environment and transferred to the apartment world by coordinate remapping, without any retraining. The Causal Circuit is constructed from the same trained JPT with the cause priority order established by ATE$_{\text{norm}}$ analysis: counter approach pose~$x$, table approach pose~$x$, arm selection, counter approach pose~$y$, and table approach pose~$y$. The effect variable is the milk's final placing height~$z$, used as a task-success proxy with threshold~$\tau$. The support determinism verification step passed before each run, providing a pre-deployment certificate that all interventional queries would be tractable. Each failed plan triggered a causal diagnosis and a one-shot corrective re-attempt; if the corrected attempt also failed, the system reverted to unconstrained JPT sampling without further correction.
 
To assess the robustness of the causal circuit across varying planning quality, two JPT variants are evaluated. The \emph{high-quality JPT} is trained on the full 1,742 samples at fine precision ($\delta{=}0.005$, $n_{\min}{=}25$), producing 53 leaves with dense coverage of the feasible approach region. The \emph{degraded JPT} is fitted with coarser precision ($\delta{=}0.15$, $n_{\min}{=}600$) and moderate approach-coordinate noise ($\sigma{=}0.18$), deliberately introducing systematic sampling failures. All other components are identical across both conditions.
 
\textbf{Results with High-Quality JPT (Efficiency).}
When the planning distribution is dense and accurate, the causal circuit has little opportunity to improve the overall success rate, as most plans already succeed on the first draw. In this condition the causal circuit provides \emph{efficiency}. As shown in Table~\ref{tab:results_good}, both the baseline JPT and the causal circuit achieve 100\% success over 5,000 iterations, confirming that the high-quality planning distribution is sufficient for this task completion in either case. The efficiency gain is instead visible in recovery cost: the causal circuit reduced total failed attempts from 97 to 87 (10.3\% reduction), average attempts per recovery from 2.04 to 1.02, and average recovery time from 24.0\,sec to 14.1\,sec, while worst-case attempts per iteration dropped from 3 to 2. Cases where $\rho_i(\hat{x}_i){=}0$ for all~$i$, meaning parameters lie entirely outside the training support, are automatically excluded from the correction loop.
 
\textbf{Results with Degraded JPT (Correctness).}
The causal circuit's role changes substantially when the planning distribution is weaker. Under the degraded JPT, coarser leaf precision causes the sampler to frequently draw parameter values near leaf boundaries that the simulator rejects, introducing systematic failures that blind resampling alone cannot reliably escape. In this condition the causal circuit provides not just efficiency but \emph{correctness}: it identifies which parameter lies outside its safe operating region and issues a targeted correction toward the high-probability region of the interventional distribution, converting failures into successes. As shown in Table~\ref{tab:results_bad}, the baseline JPT required 4,280 failed attempts across 2,283 iterations that needed resampling, averaging 2.86 attempts and 35.5 seconds per recovery, with a worst case (hard failure) of 10 attempts per iteration. The causal circuit reduced this to 2,674 failed attempts and averaging 1.13 attempts per recovery and 16.3 seconds with a 37\% reduction in wasted tester calls and a 2.2× speedup, with worst case of only 3 attempts per iteration. Under a high-quality JPT the causal circuit therefore contributes efficiency: targeted one-shot corrections consistently steer the planner toward safe parameter regions, avoiding the exhaustive blind resampling that the baseline falls into on harder iterations.

\begin{table}[t]
\centering
\caption{Performance over 5,000 iterations (high-quality JPT)}
\label{tab:results_good}
\footnotesize
\setlength{\tabcolsep}{3pt}
\renewcommand{\arraystretch}{0.7}
\begin{tabular}{@{}lcc@{}}
\hline
\textbf{Metric} & \textbf{JPT} & \textbf{JPT + Causal} \\
\hline
Successful plans               & 5,000 (100\%) & 5,000 (100\%) \\
Failed iterations              & 0             & 0             \\
Total failed attempts       & 97            & 87            \\
Corrected attempts           & 95            & 84            \\
Avg.\ attempts / recovery         & 2.04          & \textbf{1.02} \\
Avg.\ time / recovery         & 24.0          & \textbf{14.1} \\
Max attempts (one iter.)         & 3             & \textbf{2}    \\
Retry-attempt reduction percentage      & ---           & \textbf{10.3\%} \\
\hline
\end{tabular}
\end{table}

\begin{table}[t]
\centering
\caption{Performance over 5,000 iterations with the \textbf{degraded JPT}}
\label{tab:results_bad}
\footnotesize
\setlength{\tabcolsep}{3pt}
\renewcommand{\arraystretch}{0.7}
\begin{tabular}{lcc}
\hline
\textbf{Metric} & \textbf{JPT} & \textbf{JPT + Causal} \\
\hline
Successful plans          & 4,997 (99\%)  & 5,000 (100\%)  \\
Failed iterations         & 3             & 0              \\
Total failed attempts     & 4,280         & 2,674          \\
Corrected attempts        & 2,283         & 2,367          \\
Avg.\ attempts / recovery & 2.86          & \textbf{1.13}  \\
Avg.\ time / recovery     & 35.5\,sec       & \textbf{16.3\,sec} \\
Max attempts (one iter.)  & 10            & \textbf{3}     \\
Retry-attempt reduction percentage   & ---           & \textbf{37\%}  \\
\hline
\end{tabular}
\end{table}

\section{Conclusion and Future Work}
\small

We presented a closed-loop framework for causally-aware robot action testing that couples Joint Probability Trees with an Interventional Causal Circuit derived from a Marginal-Deterministic Variable Tree. When a plan fails testing, the circuit identifies the primary cause variable via the interventional distribution and issues a one-shot corrective re-attempt without retraining, additional data collection, or simulation rollouts, while a pre-deployment support determinism guarantees that all interventional queries remain tractable. On a geometrically constrained pick-and-place task over 5,000 iterations, the framework demonstrated complementary roles across planning quality: under a high-quality JPT, both conditions achieved 100\% success, with the causal circuit contributing efficiency by reducing total failed attempts by 10.3\%, average recovery time from 24.0\,sec to 14.1\,sec, and worst-case attempts per iteration from 3 to 2; under a degraded JPT, the circuit reduced total failed attempts by 37\% and recovery time by 2.2$\times$, with worst-case attempts per iteration dropping from 10 to 3. In both approaches, out-of-support plans were detected and withheld from correction automatically, ensuring the system never issues a recommendation without a grounding in observed data. Critically, every correction is interpretable: because cause variables carry explicit semantic meaning, the system produces a structured causal report naming the primary cause, its observed value, and the recommended corrective region, supporting operator oversight, audit, and safety accountability without requiring a separately trained failure model.

\textbf{Limitations.}
The current evaluation is limited to a single pick-and-place task with positional parameters; the framework has not yet been tested on complex manipulation tasks or on physical variables such as gripper force, object weight, or contact dynamics. Finally, transfer the experimentation to a real world that may test the performance of the pipeline

\textbf{Future Work.}
The most important next step is evaluating the framework on complex robot tasks, to test the full potential of the causal circuit beyond positional planning. Longer-term directions include extending the framework to sequential manipulation tasks where failure at one step propagates causally to the next, integrating the causal circuit with a continuous tester to produce soft safety scores, and validating the system on a real robot.

\normalsize
\bibliographystyle{IEEEtran}

\end{document}